%% file: sample-sigconf.tex
\documentclass[sigconf]{acmart}

\usepackage{colortbl}
\usepackage{cleveref}
\usepackage{multirow}
\usepackage{hyperref}

\definecolor{mypurple}{rgb}{0.878, 0.748, 0.996}
\definecolor{myblue}{rgb}{0.830, 0.839, 0.993}
\definecolor{mygreen}{rgb}{0.821, 0.931, 0.862}

\AtBeginDocument{%
  }

\setcopyright{cc}
\setcctype{by}
\copyrightyear{2026}
\acmYear{2026}
\acmDOI{10.1145/3770855.3817726}
\acmConference[KDD 2026]
{Proceedings of the 32nd ACM SIGKDD Conference on Knowledge Discovery and Data Mining V.2}
{August 9--13, 2026}
{Jeju Island, Republic of Korea}

\acmBooktitle{Proceedings of the 32nd ACM SIGKDD Conference on Knowledge Discovery and Data Mining V.2 (KDD 2026), August 9--13, 2026, Jeju Island, Republic of Korea}

\acmISBN{979-8-4007-2259-2/2026/08}
\begin{document}

\title{Compress the Easy, Explore the Hard: Difficulty-Aware Entropy Regularization for Efficient LLM Reasoning}

\author{Qin-Wen Luo}
\authornote{Equal Contribution. This work was performed during an internship at Didi International Business Group.}
\email{luoqw8@nuaa.edu.cn}
\affiliation{%
  \institution{Nanjing University of Aeronautics and Astronautics}
  \city{Nanjing}
  \state{Jiangsu}
  \country{China}
}

\author{Sheng Ren}
\authornotemark[1]
\email{rensheng@nuaa.edu.cn}
\affiliation{%
  \institution{Nanjing University of Aeronautics and Astronautics}
  \city{Nanjing}
  \state{Jiangsu}
  \country{China}
}

\author{Xiang Chen}
\authornote{Corresponding Author.}
\email{xiang\_chen@nuaa.edu.cn}
\affiliation{%
  \institution{College of Computer Science and Technology, Nanjing University of Aeronautics and Astronautics}
  \city{Nanjing}
  \state{Jiangsu}
  \country{China}
}

\author{Rui Liu}
\affiliation{%
  \institution{Didichuxing Co. Ltd}
  \city{Beijing}
  \country{China}
}

\author{Jun Fang}
\affiliation{%
  \institution{Didichuxing Co. Ltd}
  \city{Beijing}
  \country{China}
}

\author{Naiqiang Tan}
\affiliation{%
  \institution{Didichuxing Co. Ltd}
  \city{Beijing}
  \country{China}
}

\author{Sheng-Jun Huang}
\email{huangsj@nuaa.edu.cn}
\authornotemark[2]
\affiliation{%
  \institution{College of Computer Science and Technology, Nanjing University of Aeronautics and Astronautics}
  \city{Nanjing}
  \state{Jiangsu}
  \country{China}
}

\newcommand{\ours}{\textbf{\texttt{CEEH}}}
\renewcommand{\shortauthors}{Qin-Wen Luo et al.}

\begin{abstract}
Chain-of-Thought (CoT) has substantially empowered Large Language Models (LLMs) to tackle complex reasoning tasks, yet the verbose nature of explicit reasoning steps incurs prohibitive inference latency and computational costs, limiting real-world deployment. While existing compression methods—ranging from self-training to Reinforcement Learning (RL) with length constraints—attempt to mitigate this, they often sacrifice reasoning capability for brevity. 
We identify a critical failure mode in these approaches: explicitly optimizing for shorter trajectories triggers rapid entropy collapse, which prematurely shrinks the exploration space and stifles the discovery of valid reasoning paths, particularly for challenging questions requiring extensive deduction.
To address this issue, we propose Compress responses for Easy questions and Explore Hard ones (\textbf{\texttt{CEEH}}), a difficulty-aware approach to RL-based efficient reasoning. \textbf{\texttt{CEEH}} dynamically assesses instance difficulty to apply selective entropy regularization: it preserves a diverse search space for currently hard questions to ensure robustness, while permitting aggressive compression on easier instances where the reasoning path is well-established. In addition, we introduce a dynamic optimal-length penalty anchored to the historically shortest correct response, which effectively counteracts entropy-induced length inflation and stabilizes the reward signal.
Across six reasoning benchmarks, \textbf{\texttt{CEEH}} consistently reduces response length while maintaining accuracy comparable to the base model, and improves Pass@k relative to length-only optimization.
\end{abstract}

\ccsdesc[500]{Computing methodologies~Natural language processing}
\ccsdesc[500]{Computing methodologies~Reinforcement learning}

\keywords{Reasoning Compression; Reinforcement Learning; Entropy Regularization; Length Penalty}


\maketitle

\newcommand\kddavailabilityurl{https://doi.org/10.5281/zenodo.20375479}
\ifdefempty{\kddavailabilityurl}{}{
\begingroup\small\noindent\raggedright\textbf{Resource Availability:}\\
The source code of this paper has been made publicly available at \url{\kddavailabilityurl}. The corresponding code repository is available at
\url{https://github.com/QinwenLuo/CEEH}.
\endgroup
}

\input{section/introduction}
\input{section/preliminaries}
\input{section/method}
\input{section/experiment}
\input{section/related-work}
\input{section/conclusion}


\bibliographystyle{ACM-Reference-Format}
\bibliography{sample-base}

\input{section/appendix}

\end{document}

%% file: section/introduction.tex
\section{Introduction}
Large language models (LLMs) have achieved remarkable reliability on complex reasoning tasks by externalizing intermediate reasoning steps via Chain-of-Thought (CoT) prompting \cite{cot1, cot2, cot3, xie3}. By articulating step-by-step rationales, CoT enables models to maintain and compose intermediate evidence, thereby reducing brittle shortcut behaviors and enhancing robustness in multi-hop deduction \cite{hop1, hop2, xie1} and long-horizon planning \cite{plan1, plan2}. However, explicit reasoning often incurs significant redundancy, leading to increased inference latency, higher token consumption, and elevated serving costs. This efficiency bottleneck has motivated a growing body of research on \textit{reasoning compression} \cite{munkhbat2025self, chen2024not, fang2025when, chen2025aware}, which aims to preserve the benefits of explicit reasoning while minimizing token generation. The central challenge lies in achieving concise responses with minimal degradation in reasoning capability, effectively optimizing the trade-off between correctness and inference cost for real-world deployment.

\begin{figure}[ht]
  \begin{center}
\centerline{\includegraphics[width=0.85\columnwidth]{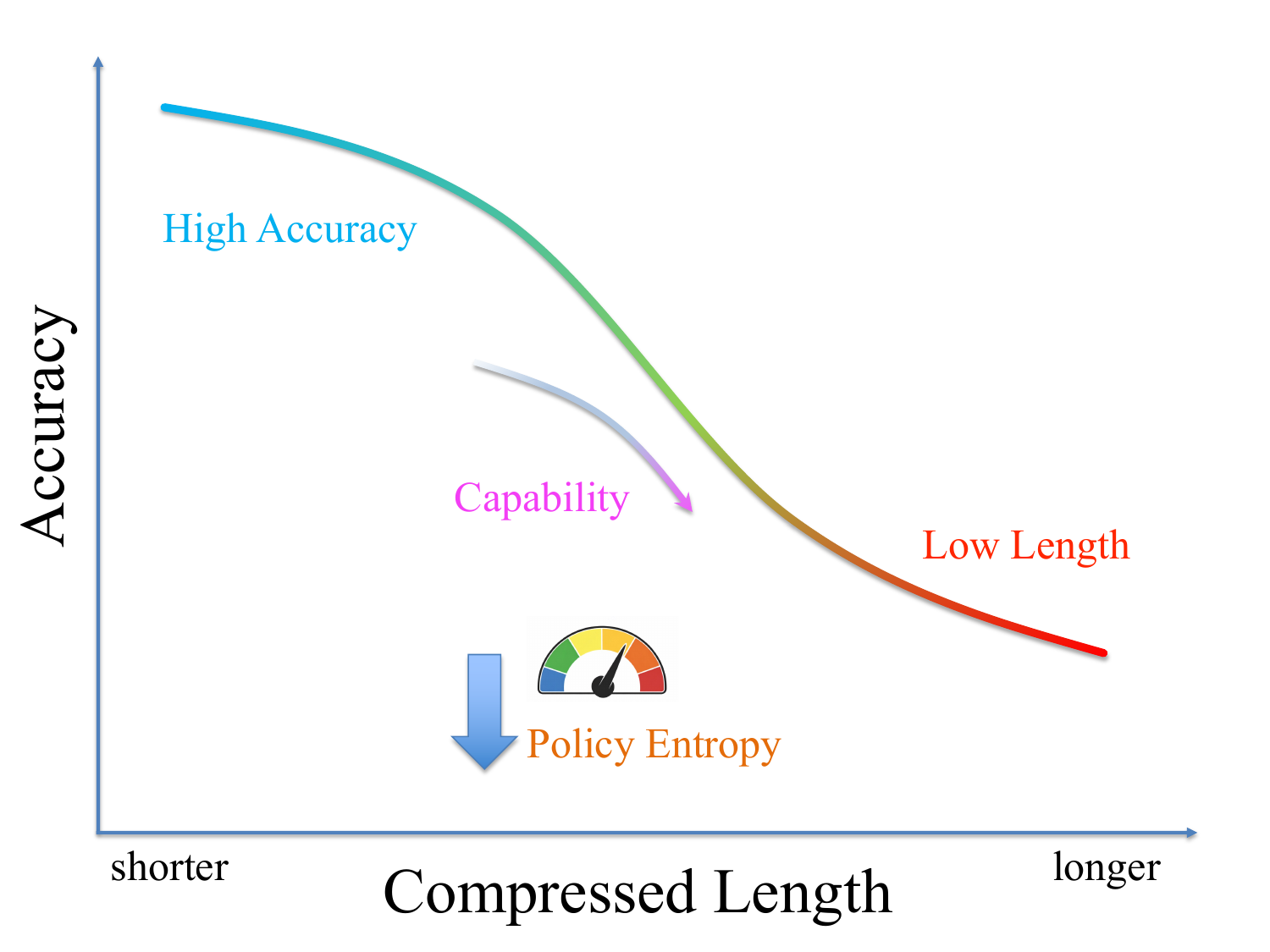}}
    \caption{Accuracy–length trade-off in reasoning compression: shorter responses can come at the cost of accuracy and reduced policy entropy.}\label{fig:trade-off}
  \end{center}
\end{figure}

A prevalent and efficient approach to compression is optimizing LLMs via RL with length-aware objectives, such as explicit length penalties \cite{arora2025length, cheng2025LC-R1, tu2025autothink}. Nevertheless, as illustrated in Figure \ref{fig:trade-off}, existing methods frequently achieve conciseness at the expense of accuracy, even when trained on high-quality datasets. We identify two primary causes for this degradation: first, overly concise generations may bypass beneficial inference behaviors like self-reflection and error correction; second, and more critically, aggressively optimizing for brevity can trigger rapid \textbf{\textit{entropy collapse}}, rendering the policy increasingly deterministic. Prior studies \cite{cui2025entropy,wang2025beyond} indicate that low policy entropy during RL impedes performance gains, as a contracted exploration space hinders the model's ability to discover correct solutions for challenging problems.

Entropy regularization \cite{cui2025entropy, wang2025beyond,park2025clip, wang2025arbitrary} offers a principled mechanism to sustain exploration during RL. However, applying uniform entropy regularization across all training instances introduces a new conflict: increased exploration typically induces longer reasoning trajectories \cite{a3po, cui2025entropy}, which directly counteracts the goal of reasoning compression and dilutes the length-control signal. This tension stems from the fact that problems of varying difficulty inherently demand different balances between exploration and compression. This observation naturally raises a pivotal question: \textbf{\textit{Which reasoning processes should be compressed, and which require thorough exploration?}}


To resolve this tension, we propose \ours, a framework grounded in the design philosophy: \textbf{C}ompress \textbf{E}asy instances and \textbf{E}xplore \textbf{H}ard ones.
The framework consists of two synergistic components.
First, we introduce \textit{difficulty-aware entropy regularization}. 
Rather than regularizing all samples uniformly, we selectively apply entropy regularization to questions currently deemed difficult for the model—instances where additional exploration is crucial to prevent premature convergence and recover accuracy. Conversely, for easier questions, regularization is relaxed, allowing the policy to confidently exploit shorter reasoning paths.  To obtain a stable difficulty signal under stochastic rollouts,we maintain per-question accuracy estimates using an asymmetric exponential moving average, effectively mitigating oscillations in difficulty classification.

To counterbalance the increased exploration on hard questions and prevent potential length inflation, we introduce a \textit{dynamic optimal-length penalty} as the second component. While entropy regularization encourages diverse reasoning, it risks undermining conventional length penalties based on fixed priors or group-wise normalization. Our dynamic penalty addresses this by tracking the historically shortest \textit{correct} response length for each question and penalizing deviations from this evolving baseline. This design ensures that exploration remains focused and efficient, providing a consistent compression signal that adapts to the model's improving capabilities.

Overall, our specific contributions can be summarized as follows:
\begin{itemize}
    \item We provide an entropy-based perspective showing that rapid entropy decay is a key factor underlying the difficulty of managing the accuracy–length trade-off in reasoning compression.
    \item We propose \ours, a novel framework that integrates \textit{difficulty-aware entropy regularization} to sustain exploration and a \textit{dynamic optimal-length penalty} to stabilize training, effectively addressing the identified collapse.
    \item Experiments on a suite of mathematical reasoning benchmarks show that {\ours} achieves a superior accuracy-efficiency trade-off compared to strong baselines, effectively compressing reasoning without sacrificing accuracy. 
\end{itemize}

%% file: section/preliminaries.tex
\section{Preliminaries}
\paragraph{Reinforcement Learning with Verifiable Rewards}
Through training on tasks with explicit, verifiable answers, Reinforcement Learning with Verifiable Rewards (RLVR) has demonstrated substantial advantages in logical reasoning domains, including mathematics, code generation, and visual question answering.
In the RLVR setting, the LLM is directly optimized as the policy $\pi_\theta$ with RL methods.
Given an input $\boldsymbol{x}$ sampled from a dataset $\mathcal{D}$, the LLM generates an output $\boldsymbol{y} = \{s, a\}$, which consists of the reasoning trajectory $s$ and a final answer $a$.
The reward of the output $\boldsymbol{y}$ is obtained by comparing the generated answer $a$ against ground-truth answer $a^{\star}$:
\begin{equation}
  R(\boldsymbol{y} | \boldsymbol{x},\pi_\theta)=
  \begin{cases}
    1, & a=a^\star \\
    0,   & a \neq a^\star
  \end{cases}
\end{equation}

A widely used algorithm for RLVR is Group Relative Policy Optimization (GRPO) \citep{shao2024deepseekmath}, which leverages group-wise advantage estimation to enable critic-free policy optimization, substantially reducing memory and computational overhead.
Specifically, for a given input $\boldsymbol{x}$, an LLM $\pi_\theta$ generates a group of $K$ outputs $\mathcal{Y} = \{\boldsymbol{y}^1,\boldsymbol{y}^2,...,\boldsymbol{y}^K\}$ and obtains the corresponding rewards $R(\boldsymbol{y}^i | \boldsymbol{x},\pi_\theta)$ by matching each answer with the ground truth.
The inter-group advantages are computed by normalizing the rewards with the group statistics:
\begin{equation}
  A(\boldsymbol{y}^i | \boldsymbol{x}) = \frac{R(\boldsymbol{y}^i | \boldsymbol{x},\pi_\theta) - \mu_R}{\sigma_R + \xi},
\end{equation}
where $\mu_R$ and $\sigma_R$ denote the mean and standard deviation of the rewards within the group, and $\xi$ is a small constant added to avoid division by zero.
Denote the importance sampling ratio $ \pi_\theta(\boldsymbol{y}^i | \boldsymbol{x}) / \pi_{\theta_{old}}(\boldsymbol{y}^i | \boldsymbol{x})$ by $\rho(\theta, \theta_{old})$. GRPO optimizes the policy model by maximizing the following objective:
\begin{equation}
  \mathcal{J}(\theta) = \mathbb{E}_{\boldsymbol{x}\sim\mathcal{D},\boldsymbol{y}^i \sim \pi_{\theta_{old}}(\boldsymbol{y}|\boldsymbol{x})} A_{\text{IS}}(\boldsymbol{y}^i|\boldsymbol{x})  - \beta \mathbb{D}_{KL}\left[\pi_{\theta} || \pi_{ref}\right],
\end{equation}
where
\begin{equation}
  A_{\text{IS}}(\boldsymbol{y}^i|\boldsymbol{x}) = \min \left[ \rho(\theta, \theta_{old}) A_i, \text{clip} \left( \rho(\theta, \theta_{old}), 1 - \epsilon, 1 + \epsilon \right)  A_i \right]
\end{equation}
and $\epsilon$ controls the trust region that constrains the policy update.

\begin{figure*}[t]
  \begin{center}
\centerline{\includegraphics[width=\textwidth]{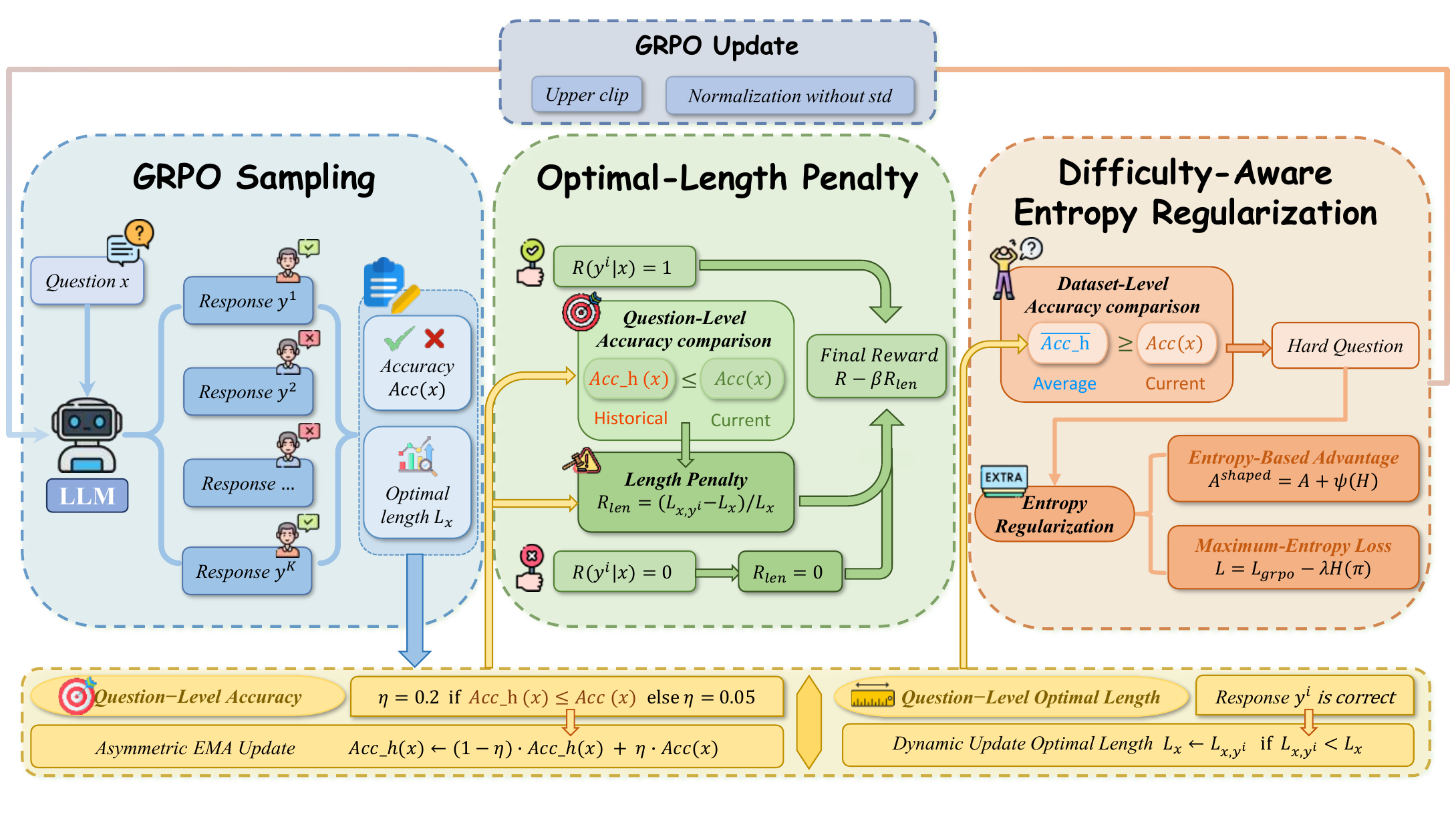}}
    \caption{The pipeline of our method. The model accuracy is evaluated via GRPO rollouts, and the optimal length is obtained from correct responses. Length penalties are applied only to correct responses when current accuracy exceeds historical accuracy, while entropy regularization is used for questions whose accuracy falls below the average to encourage exploration.}\label{fig:pipline}
  \end{center}
\end{figure*}

\paragraph{Entropy Regularization}
To encourage policy exploration, a common technique in conventional RL training is to augment the loss function with an entropy regularization term:
\begin{equation}\label{simple_entropy_regularization}
  L_{ent} = \mathbb{E}_{\boldsymbol{x} \sim D}[-\lambda_0 \mathcal{H}(\pi_\theta(\cdot|\boldsymbol{x}))],
\end{equation}
where $\mathcal{H}(\pi_\theta)$ denotes the policy entropy and $\lambda_0$ is the regularization coefficient.
For an LLM, given an input $\boldsymbol{x}$, it autoregressively generates an output sequence $\boldsymbol{y}$ that consists of $T$ tokens $\{y_1, y_2,\ldots,y_T\}$.
The policy entropy is computed as:
\begin{equation}
  \mathcal{H}(\pi_\theta(\cdot|\boldsymbol{x}))=
  \mathbb{E}_{\boldsymbol{y}\sim \pi_\theta(\cdot|\boldsymbol{x})}\left[-\frac{1}{T}\sum_{t=1}^{T}\log\pi_\theta(y_t|\boldsymbol{y}_{<t},\boldsymbol{x})\right]
\end{equation}

Recent studies suggest \cite{yue2025does} that RL does not reliably improve the Pass@$k$ performance of LLMs, and therefore may fail to yield genuine gains in reasoning capability.
Other works \cite{cui2025entropy} further observe that entropy collapse can substantially degrade Pass@$k$ performance, making entropy control during RL training a central concern.
However, naively applying Eq. \eqref{simple_entropy_regularization} as an entropy regularization term leads to strong hyperparameter sensitivity.
A small coefficient has only a minor influence on entropy, while a large one can lead to entropy explosion.
To address this issue, several studies propose LLM-specific entropy regularization techniques that achieve more stable control by identifying and training on tokens with high entropy \citep{wang2025beyond} or high covariance \citep{cui2025entropy}.
In contrast, \citet{cheng2025reasoning} introduces an entropy-based auxiliary advantage that encourages exploration by promoting longer and deeper reasoning chains, yielding a simple yet effective performance improvement.
The entropy-based advantage is defined as follows:
\begin{equation}\label{entropy-advantage}
  \psi(\mathcal{H})=\min\left(\alpha\cdot \mathcal{H}^{\mathrm{detach}},\ \frac{|A|}{\kappa}\right),
  \quad \text{where } \alpha>0 \text{ and } \kappa>1,
\end{equation}

\begin{equation}
A^{\mathrm{shaped}}=A+\psi(\mathcal{H}).
\end{equation}
Essentially, this auxiliary advantage incentivizes the generation of high-entropy tokens while allowing the reward-based advantage to remain dominant, thereby improving performance and avoiding entropy collapse.
Experiments in these studies indicate that entropy regularization tends to increase response length relative to regular RL training.

%% file: section/method.tex
\section{Methodology}

\subsection{Overview}
We present an overview of our method in Figure \ref{fig:pipline}. Our approach consists of two components.
First, to mitigate the performance degradation caused by optimizing solely for shorter reasoning length, we introduce entropy regularization to responses from high-difficulty questions in Section \ref{entropy-regularization}, promoting exploration on challenging instances and preserving overall performance.
Second, we introduce a length penalty based on the historically shortest successful response length in Section \ref{length-reward}, which provides a more informative learning signal when entropy regularization possibly induces longer outputs.
Finally, Section \ref{Implementation} details some important implementation during training.

\subsection{Difficulty-Aware Entropy Regularization}\label{entropy-regularization}

\paragraph{Motivation.}
Standard RL-based reasoning compression methods \cite{cheng2025LC-R1, arora2025length} typically rely on length penalties to incentivize brevity. However, aggressive length optimization often precipitates a rapid collapse in policy entropy, narrowing the exploration space and causing the model to become overly confident in suboptimal, shortcut solutions. This is particularly detrimental for complex reasoning tasks, where maintaining diverse reasoning paths is essential for error correction and self-reflection \citep{guo2025deepseek}. Furthermore, prior work suggests that reasoning-critical tokens—those governing logical structure—inherently exhibit high entropy \citep{hou2025treerl, wang2025beyond}. Consequently, suppressing entropy indiscriminately risks degrading the model's logical deduction capabilities. To address this, we propose \textit{difficulty-aware entropy regularization}, a strategy that selectively sustains exploration on hard instances while permitting aggressive compression on easier ones.

\paragraph{Dynamic Difficulty Estimation.}
To implement this strategy, we first need a robust metric to distinguish ``hard'' questions from ``easy'' ones during training. It is intractable to estimate the accurate difficulty from limited stochastic rollouts in GRPO. Therefore, we maintain a historical accuracy estimate for each question using an asymmetric Exponential Moving Average (EMA).
Specifically, for a question $\boldsymbol{x}$, we sample $K$ responses in each step and compute the instantaneous accuracy:
\begin{equation}
  \mathrm{Acc}(\boldsymbol{x}) = \frac{1}{K}\sum_{i=1}^{K}\mathbb{I}\left[a^{i}=a^{\star}\right].
\end{equation}
We then update a running historical accuracy score, $\mathrm{Acc}_h(\boldsymbol{x})$, which serves as a stable proxy for the model's mastery of the specific problem (implementation details in Section \ref{Implementation}).
A question is classified as \textit{high-difficulty} ($\mathcal{D}_h$) if its historical accuracy falls below the global average accuracy of the entire dataset:
\begin{equation}
  \mathcal{D}_h = \left\{\boldsymbol{x} \in \mathcal{D} \;\middle|\; \mathrm{Acc}_h(\boldsymbol{x}) < \frac{1}{|\mathcal{D}|}\sum_{\boldsymbol{x}^\prime \in \mathcal{D}}\mathrm{Acc}_h(\boldsymbol{x}^\prime)\right\}.
\end{equation}
This dynamic thresholding ensures that the definition of ``difficulty'' evolves as the model improves.

\paragraph{Selective Regularization Mechanism.}
Based on the difficulty classification, we apply entropy regularization selectively to prevent premature convergence on hard problems. 
We employed two distinct forms of entropy regularization, respectively.

First, following standard RL practice, we augment the training objective with the entropy-maximization term in Eq. \ref{simple_entropy_regularization}.
We use a cosine annealing schedule for the coefficient $\lambda(t)$, but crucially, we amplify the regularization strength for hard questions. The coefficient schedule is defined as:
\begin{equation}
  \lambda(\boldsymbol{x}, t) =
  \begin{cases}
    5 \cdot \lambda_0 \cdot \cos \left(\frac{\pi t}{T}\right), & \boldsymbol{x} \in \mathcal{D}_h \\
    \lambda_0 \cdot \cos \left(\frac{\pi t}{T}\right), & \boldsymbol{x} \in \mathcal{D} \setminus \mathcal{D}_h
  \end{cases}
\end{equation}
where $t$ is the current step and $T$ is the total training steps. The $5\times$ multiplier forces the policy to maintain a wider exploration frontier for challenging questions.

Second, we incorporate an entropy-based advantage term $\psi(\mathcal{H})$ defined in Eq. \ref{entropy-advantage} (following \cite{wang2025beyond}) exclusively for the hard subset. The shaped advantage function becomes:
\begin{equation}
  A^{\mathrm{shaped}}(\boldsymbol{y}^i|\boldsymbol{x}, \pi_\theta)=
  \begin{cases}
    A(\boldsymbol{y}^i|\boldsymbol{x}, \pi_\theta) + \psi(\mathcal{H}), & \boldsymbol{x} \in \mathcal{D}_h \\
    A(\boldsymbol{y}^i|\boldsymbol{x}, \pi_\theta), & \boldsymbol{x} \in \mathcal{D} \setminus \mathcal{D}_h
  \end{cases}
\end{equation}
With selective and difficulty-aware entropy regularization, the
model sustains adequate exploration on high-difficulty problems,
mitigating the performance drop that can arise when response-length
reduction leads to insufficient reasoning.

\input{section/main_table}

\subsection{Dynamic Optimal-Length Penalty}\label{length-reward}
Prior length-penalty methods are mainly based on either discouraging deviations from a predefined target length or applying penalties through inter-group length normalization and comparison.
The former relies on hand-specified priors that often do not transfer across datasets or tasks.
The latter can behave poorly when the average response length of the group is large, since it may assign positive advantage to overly long outputs, thereby weakening the compression signal and reducing training efficiency.
These issues are further amplified when entropy regularization is used: to maintain accuracy on difficult questions, the model may produce longer responses.
In addition, because response lengths often decrease substantially over the course of training, the penalties based on the target length can become highly non-stationary across phases, which typically requires additional tuning of the penalty coefficient.

To address these limitations, we propose a question-level length penalty based on a dynamically updated optimal length.
Specifically, for a question $\boldsymbol{x}$, we track the shortest length among all historically correct responses, denoted by $L_{\boldsymbol{x}}$, and update it online throughout training.
To encourage the model to produce correct solutions with the minimum amount of tokens, we only penalize the length of the correct responses. We define the length penalty as:
\begin{equation}
  R_{len}(\boldsymbol{y}^i|\boldsymbol{x},\pi_\theta)=\frac{L_{\boldsymbol{x},\boldsymbol{y}^i}-L_{\boldsymbol{x}}}{L_{\boldsymbol{x}}} \cdot \mathbb{I}\left[a^{i}=a^{\star}\right].
\end{equation}
Where $\mathbb{I}(\cdot)$ is the indicator function.
Meanwhile, we apply length penalties at the question level by enabling the penalty only when the current accuracy for the given question exceeds its historical estimate, i.e., $\mathrm{Acc}(\boldsymbol{x}) > \mathrm{Acc}_h(\boldsymbol{x})$.
This design enables the model to compress its reasoning process while preserving its original performance.
To avoid reversing the reward relationship between correct and incorrect answers, we clip the minimum value of the length term to $-0.9$, so that the length penalty lies in the range $[-0.9, 1)$.
The total reward used for RL training is:
\begin{equation}
  R_{total}(\boldsymbol{y}^i| \boldsymbol{x},\pi_\theta) = R(\boldsymbol{y}^i | \boldsymbol{x},\pi_\theta) - \beta R_{len}(\boldsymbol{y}^i|\boldsymbol{x},\pi_\theta),
\end{equation}
where $\beta$ controls the strength of the length term.

Since the question-level optimal length typically decreases over training in tandem with the overall reduction in average response length, it provides a stable and well-calibrated penalty signal across training phases.
Moreover, when entropy regularization causes response lengths to increase for certain questions, the shorter optimal length as the denominator yields a more discriminative penalty, strengthening the compression signal precisely when length inflation occurs.

\subsection{Implementation}\label{Implementation}
We include several implementation details that are important for reproducing our results.

Firstly, to match the work of entropy-based advantage \citep{cheng2025reasoning}, we adopt the Clip-Higher strategy by increasing the upper clipping threshold to 0.28, which empirically promotes policy exploration.
DAPO \citep{yu2025dapo} first adopts this technique to relax the constraint on increasing the probabilities of low-probability exploratory tokens, thereby encouraging broader exploration.

Secondly, we update the per-question accuracy estimates using an asymmetric exponential moving average (EMA):
\begin{equation}
  \mathrm{Acc_h}
  (\boldsymbol{x}) \leftarrow (1-\eta_{\boldsymbol{x}})\,\mathrm{Acc_h}(\boldsymbol{x})+\eta_{\boldsymbol{x}}\,\mathrm{Acc}(\boldsymbol{x}),
\end{equation}
where the update rate $\eta_{\boldsymbol{x}}$ is defined as
\begin{equation}
  \eta_{\boldsymbol{x}}=
  \begin{cases}
    0.2, & \mathrm{Acc_h}(\boldsymbol{x}) < \mathrm{Acc}(\boldsymbol{x})\\
    0.05, & \mathrm{Acc_h}(\boldsymbol{x}) \ge \mathrm{Acc}(\boldsymbol{x})
  \end{cases}
\end{equation}
and $\mathrm{Acc}(\boldsymbol{x},t)$ denotes the accuracy of question $\boldsymbol{x}$ at the current training step $t$.
This smooths the inherently noisy accuracy measurements arising from stochastic sampling and finite $K$ rollouts, yielding a more stable difficulty signal for selective entropy regularization and reducing oscillations in the set of questions classified as hard ones across training steps.

Finally, following \citep{arora2025length}, we do not apply advantage standardization; instead, we only subtract a reward-mean baseline. This avoids the scale distortion introduced by normalizing with the reward standard deviation, which can inadvertently attenuate the effect of the length term and make the length-penalty coefficient less effective for controlling response length.

%% file: section/main_table.tex
\begin{table*}[t!]
  \centering
    \caption{Performance comparison across mathematical benchmarks. \textbf{Bold} indicates the best score for each metric, and the \textcolor{gray}{accuracy} in grey denotes the second-best score. Here, ``\raisebox{1pt}{\colorbox{mypurple}{ \rule[-0.2ex]{0pt}{1.0ex} }}'': prompting methods, ``\raisebox{1pt}{\colorbox{myblue}{ \rule[-0.2ex]{0pt}{1.0ex} }}'': offline methods, ``\raisebox{1pt}{\colorbox{mygreen}{ \rule[-0.2ex]{0pt}{1.0ex} }}'': online methods. ``$*$'' indicates results reproduced or re-evaluated in this study.}
  \resizebox{\textwidth}{!}{
    \begin{tabular}{l | ccc | ccc | ccc | ccc | ccc | ccc}
      \toprule
      \textbf{Model Name} & \multicolumn{3}{c|}{\textbf{GSM8K}} & \multicolumn{3}{c|}{\textbf{MATH500}} & \multicolumn{3}{c|}{\textbf{AIME24}} & \multicolumn{3}{c|}{\textbf{AMC}} & \multicolumn{3}{c|}{\textbf{OlymBench}} & \multicolumn{3}{c}{\textbf{AIME25}} \\
      \cmidrule(lr){2-19}
      & ACC & LEN & NAG $\downarrow $ & ACC & LEN & NAG $\downarrow $ & ACC & LEN & NAG $\downarrow $ & ACC & LEN & NAG $\downarrow $ & ACC & LEN & NAG $\downarrow $ & ACC & LEN & NAG $\downarrow $ \\
      \midrule
      Qwen2.5-7B-Ins & 90.9 & 279.0 & 0.3 & 74.2 & 567.0 & 17.24 & 12.0 & 1016.0 & 72.46 & 47.5 & 801.0 & 42.0 & 39.2 & 827.0 & 37.85 & 7.6 & 1240.0 & 74.67 \\
      Qwen2.5-7B-Math & 93.2 & 439.0 & -1.84 & 63.4 & 740.0 & 27.33 & 19.0 & 1429.0 & 57.99 & 62.5 & 1022.0 & 25.41 & 31.5 & 1037.0 & 48.14 & 4.0 & 2562.0 & 77.99 \\
      Qwen2.5-7B-Math-Ins & 95.2 & 323.0 & -3.88 & 81.4 & 670.0 & 9.81 & 10.3 & 1363.0 & 74.23 & 60.0 & 1029.0 & 27.99 & 38.9 & 1027.0 & 37.69 & 9.3 & 2087.0 & 67.17 \\
      \midrule
      R1-Distill-Qwen2.5-7B* & 91.2 & 1479 & -- & 91.3 & 3701 & -- & 50.6 & 10382 & -- & 86.9 & 5646 & & 65.5 & 7413 & -- & 36.7 & 10958 & -- \\
      \rowcolor[rgb]{0.942, 0.881, 0.998}
      \ \ + ThinkSwitcher & \textbf{92.5} & 1389.0 & \textcolor{gray}{-0.35} & 91.3 & 3495.0 & 0.0 & 48.3 & 7936.0 & 2.21 & -- & -- & -- & 57.0 & 5147.0 & 7.17 & \textbf{37.5} & \textbf{6955} & \textbf{-1.32} \\
      \rowcolor[rgb]{0.942, 0.881, 0.998}
      \ \ + Dynasor-CoT & 89.6 & 1285.0 & 0.64 & 89.4 & 2661.0 & 1.1 & 46.7 & 12695.0 & \ & 85.0 & 5980.0 & \ & -- & -- & -- & -- & -- & -- \\
      \rowcolor[rgb]{0.942, 0.881, 0.998} \ \ + DEER & 90.6 & 917.0 & 0.41 & 89.8 & 2143.0 & 1.07 & 49.2 & 9839.0 & 0.63 & 85.0 & 4451.0 & 1.01 & -- & -- & -- & -- & -- & -- \\
      \rowcolor[rgb]{0.928, 0.936, 0.997} \ \ + Spirit & 87.2 & 687.0 & 3.21 & 90.8 & \textcolor{gray}{1765} & 0.4 & 38.3 & \textcolor{gray}{6926} & 14.02 & -- & -- & -- & -- & -- & -- & -- & -- & -- \\
      \rowcolor[rgb]{0.928, 0.936, 0.997} \ \ + ConCISE-SimPO & \textcolor{gray}{92.1} & 715.0 & \textbf{-0.71} & 91.0 & 1945.0 & 0.23 & 48.3 & 7745.0 & 2.29 & -- & -- & -- & -- & -- & -- & -- & -- & -- \\
      \rowcolor[rgb]{0.928, 0.936, 0.997} \ \ + DAST & 86.7 & \textcolor{gray}{459} & 4.1 & 89.6 & 2162.0 & 1.2 & 45.6 & 7578.0 & 5.14 & -- & -- & -- & -- & -- & -- & -- & -- & -- \\

      \rowcolor[rgb]{0.919, 0.969, 0.938} \ \ + AutoThink & -- & -- & -- & 91.2 & 2146.0 & 0.07 & \textbf{54.8} & 8051.0 & \textbf{-3.93} & 83.3 & 4645.0 & 1.74 & 56.4 & 5498.0 & 7.06 & -- & -- & -- \\

      \rowcolor[rgb]{0.919, 0.969, 0.938} \ \ + LC-R1 & 88.1 & \textbf{450} & 2.84 & 90.4 & \textbf{1568} & 0.75 & -- & -- & -- & 79.1 & \textbf{3453} & 5.59 & 58.7 & \textbf{4041} & 7.0 & 36.2 & \textcolor{gray}{7150} & 0.8 \\

      \rowcolor[rgb]{0.919, 0.969, 0.938} \ \ + Length-Penalty* & 91.6 & 931.0 & -0.27 & 91.4 & 2696.0 & -0.06 & 50.2 & 9517.0 & 0.23 & 85.0 & 4937.0 & 0.77 & 64.8 & 6411.0 & 0.39 & 35.6 & 10173.0 & 0.8 \\

      \rowcolor[rgb]{0.919, 0.969, 0.938} \ \ + \textbf{{\ours}-EA}  & 91.3 & 723.0 & -0.08 & \textcolor{gray}{91.7} & 2327.0 & \textcolor{gray}{-0.27} & 53.5 & 7543.0 & -3.0 & \textbf{88.1} & 4014.0 & \textbf{-0.74} & \textcolor{gray}{66.3} & 5253.0 & \textcolor{gray}{-0.66} & \textcolor{gray}{37.1} & 8327.0 & \textcolor{gray}{-0.53} \\

      \rowcolor[rgb]{0.919, 0.969, 0.938} \ \ + \textbf{{\ours}-ME} & 91.3 & 646.0 & -0.08 & \textbf{92.1} & 2170.0 & \textbf{-0.56} & \textcolor{gray}{53.8} & \textbf{6824} & \textcolor{gray}{-3.7} & \textcolor{gray}{87.3} & \textcolor{gray}{3474} & \textcolor{gray}{-0.29} & \textbf{66.9} & \textcolor{gray}{4383} & \textbf{-1.37} & 36.3 & 7311.0 & 0.63 \\

      \midrule
      R1-Distill-Qwen2.5-1.5B* & 84 & 1879 & -- & 81.5 & 4772 & -- & 27.7 & 12164 & -- & 65.5 & 8127 & -- & 41.5 & 8957 & -- & 21.7 & 11981 & -- \\
      \rowcolor[rgb]{0.919, 0.969, 0.938} \ \ + AutoThink & -- & -- & -- & \textbf{84} & \textcolor{gray}{2195} & \textbf{-2.25} & \textbf{34.6} & 9514.0 & \textbf{-11.63} & 67.0 & 5059.0 & -1.41 & \textbf{44.8} & 5559.0 & \textcolor{gray}{-4.9} & -- & -- & -- \\
      \rowcolor[rgb]{0.919, 0.969, 0.938} \ \ + LC-R1 & 80.2 & \textbf{621} & 3.7 & 79.3 & \textbf{1822} & 2.12 & -- & -- & -- & 59.0 & \textbf{3591} & 7.41 & 42.7 & \textbf{3780} & -2.2 & 20.8 & \textbf{5953} & 2.94 \\
      \rowcolor[rgb]{0.919, 0.969, 0.938} \ \ + \textbf{{\ours}-EA} & \textbf{85.8} & \textcolor{gray}{997} & \textbf{-1.47} & 83.6 & 2497.0 & \textcolor{gray}{-1.78} & \textcolor{gray}{29.8} & \textbf{7713} & \textcolor{gray}{-4.59} & \textcolor{gray}{72.3} & \textcolor{gray}{4044} & \textbf{-7.36} & \textcolor{gray}{44.7} & \textcolor{gray}{4693} & \textbf{-5.32} & \textbf{24} & \textcolor{gray}{7350} & \textbf{-6.59} \\
      \rowcolor[rgb]{0.919, 0.969, 0.938} \ \ + \textbf{{\ours}-ME} & \textcolor{gray}{85.6} & 1001.0 & \textcolor{gray}{-1.3} & \textcolor{gray}{83.8} & 3067.0 & -1.69 & 28.5 & \textcolor{gray}{8511} & -1.58 & \textbf{72.6} & 5078.0 & \textcolor{gray}{-6.64} & 44.3 & 5469.0 & -4.21 & \textcolor{gray}{23.1} & 8390.0 & \textcolor{gray}{-3.53} \\

      \bottomrule
    \end{tabular}
  }
  \label{tab:main_results}
\end{table*}

%% file: section/experiment.tex
\section{Experiments}
\subsection{Setup}
\paragraph{Datasets and Models}
We adopt the verl framework \citep{sheng2024hybridflow} for reinforcement learning training.
We randomly sample 2500 instances from each of the DeepMath103K \citep{he2025deepmath} and DAPO \citep{yu2025dapo} datasets, merge them to construct the training set, and use R1-distill-\allowbreak Qwen-2.5-\allowbreak 7B \citep{guo2025deepseek} as the base model.
The trained model is evaluated on GSM8K \citep{cobbe2021gsm8k}, Math-500 \citep{he2025deepmath}, AIME24, AIME25, AMC23, and OlympiadBench \citep{he2024olympiadbench}.
To mitigate the impact of sampling randomness, we set the sampling temperature to 0.6 and perform 16 independent rollouts, reporting the average accuracy avg@16.
Similarly, we report the average number of generated tokens to demonstrate the effectiveness of our method in compressing reasoning.

\begin{figure}[ht!]
  \begin{center}
    \centerline{\includegraphics[width=\columnwidth]{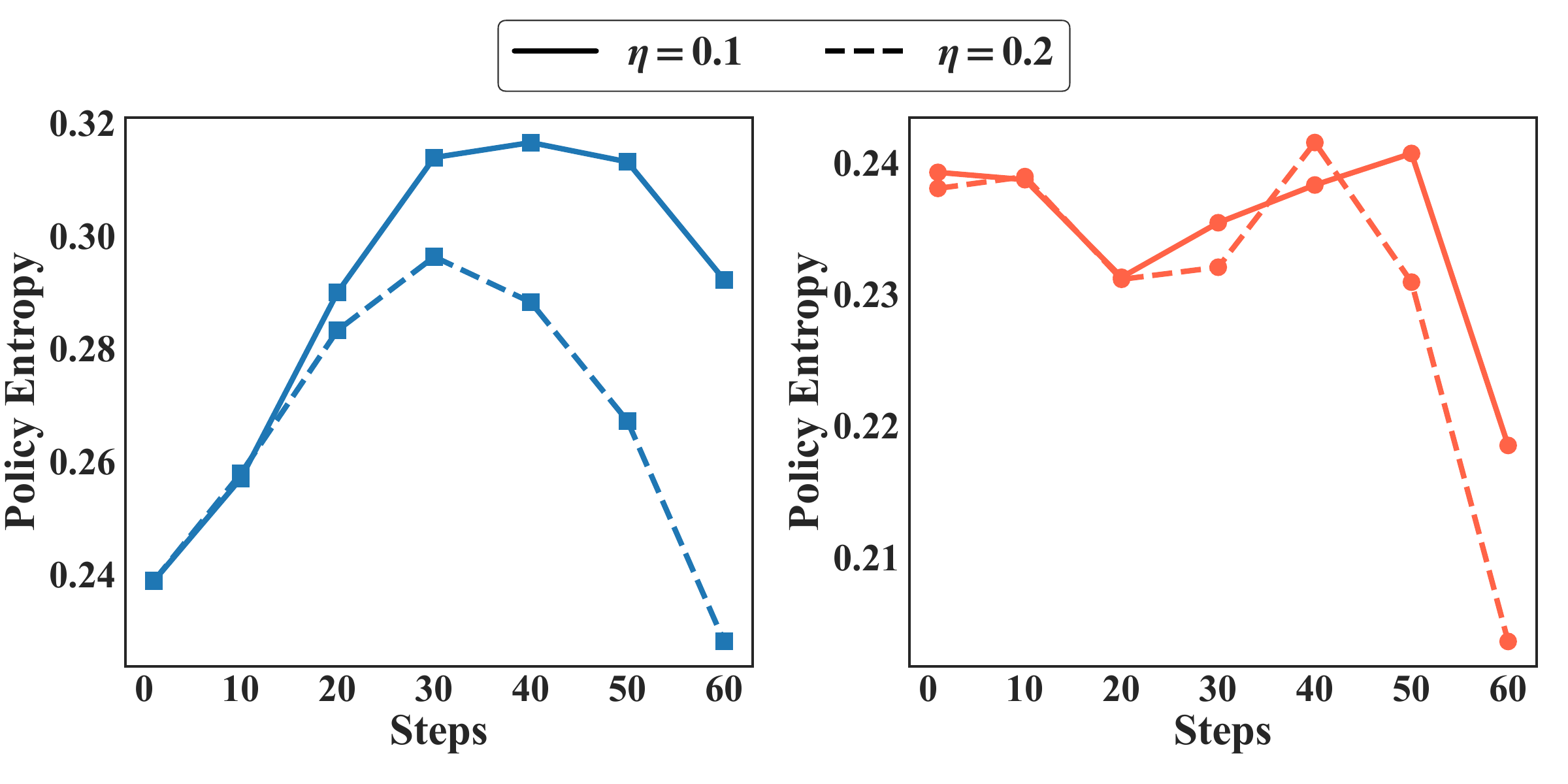}}
    \caption{Training dynamics of policy entropy for R1-Distill-Qwen2.5-7B under different length penalty coefficients. (Left: Maximum-entropy loss ({\ours}-ME); Right: Entropy-based advantage ({\ours}-EA)).
    }
    \label{fig:entropy_comparison}
  \end{center}
\end{figure}

\paragraph{Baseline}
We consider three representative paradigms as baselines for a comprehensive comparison:
(1) \textbf{Prompting-based methods}: ThinkSwitcher \citep{ThinkSwitcher} trained  a lightweight switching module with supervision signals to dynamically switch between short
and long CoT modes based on task complexity; Dynasor-CoT \citep{fu2025reasoning} and DEER \citep{DEER} propose approaches for dynamic reasoning termination.
(2) \textbf{Offline methods}: Spirit \citep{Spirit} and ConCISE-SimPO \citep{ConCISE-SimPO} utilize self-generated responses to compress reasoning via confidence scores or preference optimization; DAST \citep{DAST} introduces a difficulty metric and applies budget-aware reward shaping together with budget preference optimization.
(3) \textbf{Online RL methods}: AutoThink \citep{tu2025autothink} progressively refines reasoning strategies through staged reward shaping, achieving favorable accuracy–efficiency
trade-offs; LC-R1 \citep{cheng2025LC-R1} compresses reasoning by sampling target lengths for each question and applying length-based rewards to encourage generations that match these targets; Length-Penalty \citep{arora2025length} assigns inter-group length rewards using normalized response lengths, providing a richer length-control signal for optimizing generation length. We report results from the original papers or subsequent evaluations when available, and we additionally reproduce Length-Penalty under the same training data used in our experiments.
We denote \textbf{\ours-EA} as using the entropy-based advantage for entropy regularization, and \textbf{\ours-ME} as using the maximum-entropy loss for entropy regularization.
In addition, we also included some instruction models in the comparison, such as Qwen2.5-Math \cite{Qwen}.

\begin{table}[ht!]
    \caption{Pass@$k$ performance of different methods trained from R1-Distill-Qwen2.5-7B. Pass@$k$ is computed using 16 rollouts per question.}
  \centering
  \resizebox{\columnwidth}{!}{
    \begin{tabular}{l | c | c | c | c | c | c}
      \toprule
      \textbf{Model Name} & \textbf{GSM8K} & \textbf{MATH500} & \textbf{AIME24} & \textbf{AMC} & \textbf{OlymBench} & \textbf{AIME25} \\
      \midrule
      Base Model & 97.8 & 97.2 & 80 & 97.5 & 81.8 & 63.3 \\

      Length-Penalty & 97.6 & 97 & 76.7 & 97.5 & 81.8 & 63.3 \\

      \textbf{{\ours}-EA} & 98.1 & 97.2 & 80 & 97.5 & 82.4 & 63.3 \\
      \textbf{{\ours}-ME} & 98.3 & 97.2 & 80 & 97.5 & 82.2 & 70 \\

      \bottomrule
    \end{tabular}
  }
  \label{tab:passk}
\end{table}

\paragraph{Metrics}
We report the accuracy with avg@16 and the tokens of responses, denoted as ACC and LEN.
Prior work often evaluates reasoning compression primarily through length-based metrics.
Since our goal is to compress reasoning length without sacrificing task performance, we introduce a metric that jointly accounts for length and accuracy relative to the base model.
Following prior work \cite{Bingo}, we define \textbf{Normalized Accuracy Gain (NAG)} as:
\begin{equation}
  \text{NAG} = (1 - Acc/Acc_b) \times 100 /  \sqrt{1-L/L_b}
\end{equation}
where $Acc_b$ and $L_b$ denote the accuracy and average response length of the base model. If the response length $L$ is larger than the reference length $L_b$, we do not compute this metric since it does not compress the reasoning progress.
Intuitively, this metric quantifies how much accuracy is sacrificed per unit reduction in response length, balancing performance degradation against efficiency gains.
Compared to other indicators, this metric emphasizes changes in relative proportions rather than absolute magnitudes.

\subsection{Main Results}

\textbf{\ours~ can compress reasoning without sacrificing model accuracy.}
Table \ref{tab:main_results} summarizes performance and token efficiency across six reasoning benchmarks.
Overall, {\ours} provides a more reliable trade-off between reasoning compression and accuracy preservation than prompting-based, offline, or online optimization baselines.
In particular, both entropy regularization variants consistently and substantially reduce response length, while maintaining accuracy comparable to or even better than that of the base model.
This indicates that {\ours} can compress reasoning without systematically sacrificing correctness.

\begin{figure}[htp!]
  \begin{center}
\centerline{\includegraphics[width=0.95\columnwidth]{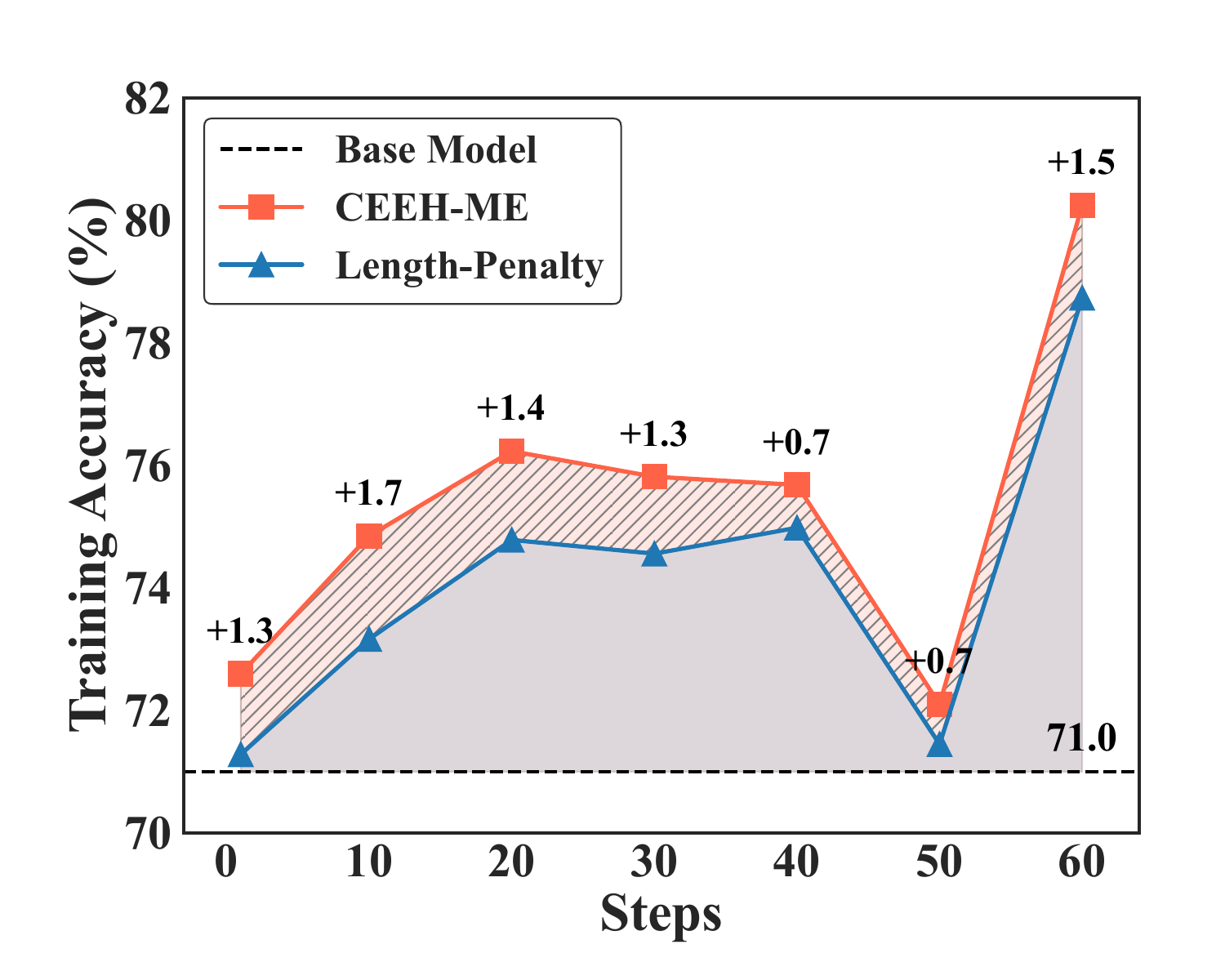}}
    \caption{Training accuracy on the same dataset, with R1-Distill-Qwen2.5-7B as the base model.}
\label{fig:training_acc_comparison}
  \end{center}
\end{figure}

\textbf{\ours~ is more robust across tasks.} The compression effect of {\ours} generalizes across tasks with diverse difficulties and reasoning styles, rather than being benchmark-specific: both entropy-regularization variants reduce total response tokens by over 30\%.
Prompting-based and offline methods can shorten responses on select benchmarks, but their accuracy tends to fluctuate more.
Among online methods, Length-Penalty exhibits substantial variability in token reduction across datasets, while AutoThink shows considerable discrepancies in accuracy across benchmarks.
In addition, {\ours} achieves strong compression while also attaining the best or the second-best accuracy on several benchmarks.
These results demonstrate that efficiency gains need not come at the expense of accuracy when exploration is appropriately controlled.

\textbf{\ours~ achieves better performance improvements on stronger models.}
Table \ref{tab:main_results} shows that {\ours} attains a better trade-off between accuracy and response length on R1-Distill-Qwen2.5-7B.
When trained on R1-Distill-Qwen2.5-1.5B, LC-R1 generates more concise responses but falls below the base model in accuracy. In contrast, AutoThink achieves shorter responses while even improving accuracy.
One plausible explanation is that weaker models have more headroom to benefit from RL fine-tuning, whereas stronger models are harder to improve in accuracy.
Moreover, their higher baseline accuracy yields denser length-related reward signals, which encourages the policy to focus more on compressing reasoning.
As a result, many methods can further shorten responses only by trading off correctness. Although {\ours} is not the top performer on some R1-Distill-Qwen2.5-1.5B benchmarks, it consistently reduces response length while preserving accuracy across model scales.

\subsection{Feature Analysis of Training Dynamics}
In this subsection, we analyze the training dynamics to understand when and why our method is effective, focusing on policy-entropy evolution, Pass@$k$ performance, and training accuracy trends.
\paragraph{Dynamics of Policy Entropy}

\textbf{How does the model’s policy entropy evolve during training?}
Revisiting our motivation, we aim for the model’s policy entropy to increase on difficult problems to ensure sufficient exploration.
Figure \ref{fig:entropy_comparison} illustrates the evolution of entropy during training under different entropy regularization terms and length penalty coefficients.
When using the maximum-entropy loss, the additional entropy regularization leads to an increase in policy entropy. Consequently, the policy entropy rises in the early stage of training. However, since we apply cosine annealing to the entropy regularization coefficient, the policy entropy decreases in the later stage of training. Applying entropy regularization only to difficult problems helps avoid entropy explosion \cite{cui2025entropy}, keeping the overall entropy under control.
While using entropy-based advantages, the method inherently encourages the generation of high-entropy tokens while keeping reward advantages dominant, enabling adaptive entropy adjustment. As a result, the model’s policy entropy remains relatively stable in the early training stage. In the later stage, as accuracy improves, the contribution of entropy-based advantages becomes smaller, and the entropy gradually decreases.

\begin{figure}[ht]
  \begin{center}
\centerline{\includegraphics[width=1.1\columnwidth]{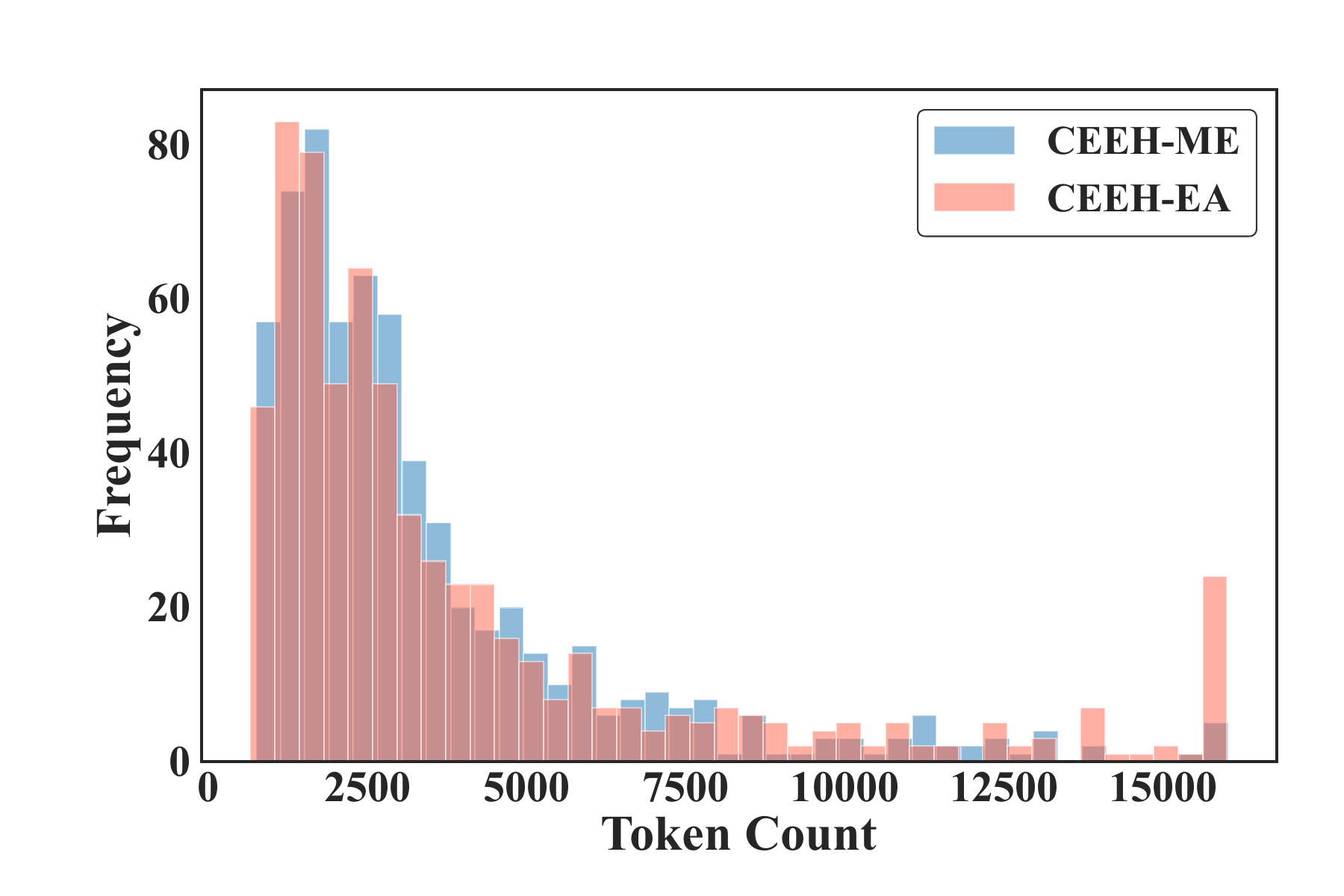}}
    \caption{Distribution of response token counts on AMC23, with R1-Distill-Qwen2.5-7B as the base model.}
    \label{fig:token_distribution_comparison}
  \end{center}
\end{figure}

\paragraph{Impact of the Length-Penalty Coefficient on Policy Entropy}
\textbf{How does the length penalty coefficient affect policy entropy?} Figure \ref{fig:entropy_comparison} shows that, for both entropy regularization variants, increasing the length penalty coefficient consistently reduces policy entropy.
Intuitively, stronger constraint toward shorter outputs narrows the model’s action distribution, encouraging more deterministic token choices.
This effect is particularly pronounced under on-policy RL, where updates reinforce the model’s current generations with high confidence.
This trend supports our claim that optimizing for length alone accelerates entropy collapse, limits exploration, and can ultimately result in accuracy sacrifice.

\paragraph{Pass@k Performance}

\textbf{Can \ours~ improves the model’s underlying reasoning capability?} Prior work \cite{chen2025pass, cheng2025reasoning} highlights policy entropy as a key factor governing Pass@$k$ performance, which is commonly considered as an upper-bound estimator for the true reasoning capabilities of an LLM.
Motivated by this connection, we analyze why our difficulty-aware entropy regularization can preserve accuracy while compressing reasoning, from the perspective of Pass@$k$.
Table \ref{tab:passk} compares Pass@$k$ for {\ours} against the base model and the Length-Penalty baseline. Notably, Length-Penalty consistently degrades Pass@$k$ by optimizing solely for reasoning compression, even on the AIME datasets where the reduction in response length is relatively modest.
In contrast, {\ours} maintains adequate exploration on challenging instances, leading to improved Pass@$k$ and an avg@16 gain. These results suggest that {\ours} enhances inference ability via reasonable entropy control, rather than merely increasing the model’s confidence in its own predictions.

\begin{figure}[htp!]
  \begin{center}    \centerline{\includegraphics[width=\columnwidth]{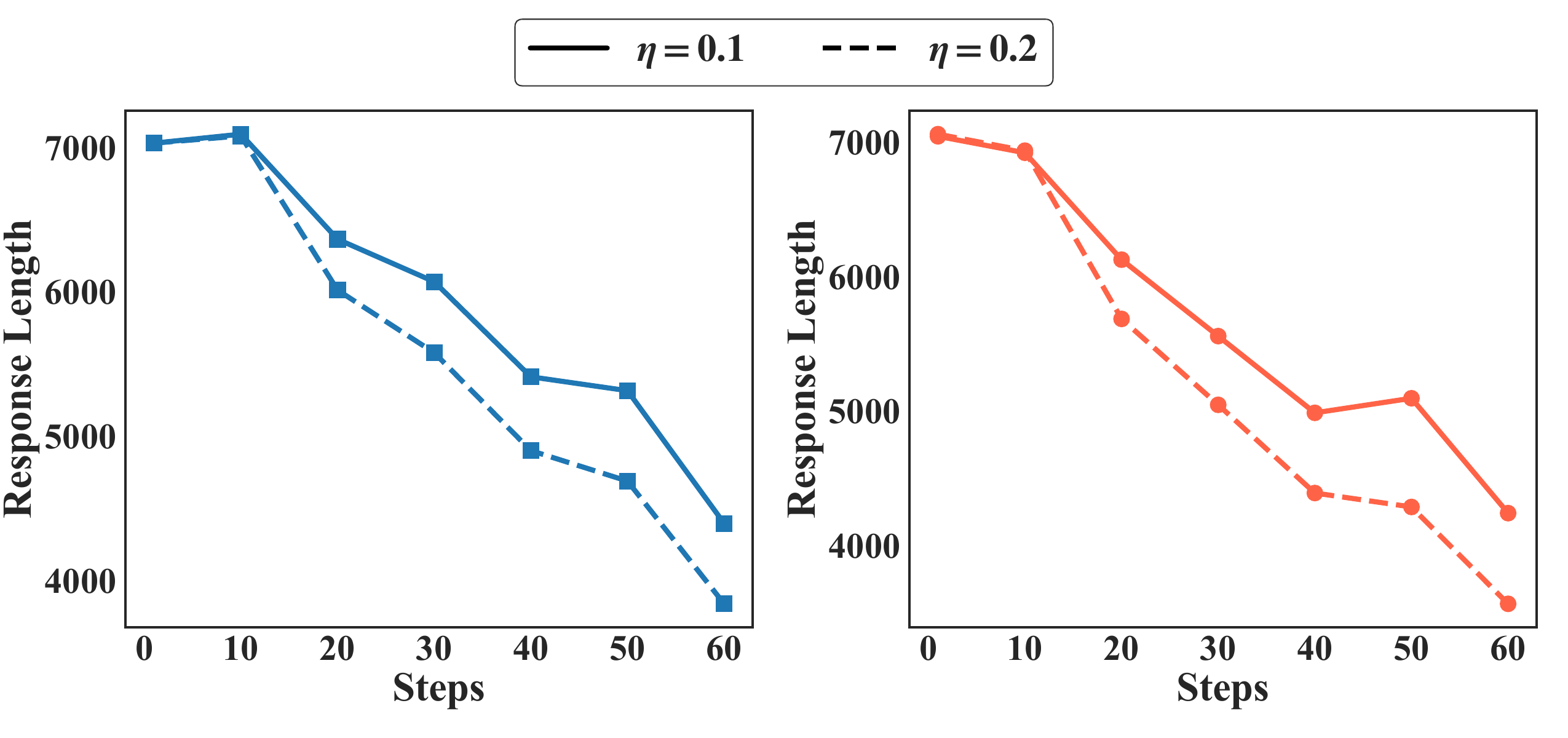}}
    \caption{The dynamics of response length during training with R1-Distill-Qwen2.5-7B as the base model. (Left: Maximum-entropy loss ({\ours}-ME); Right: Entropy-based advantage ({\ours}-EA).}
    \label{fig:length_comparison}
  \end{center}
\end{figure}

\begin{table}[ht]
  \caption{Performance under different length penalty coefficients.}
  \centering
  \resizebox{\columnwidth}{!}{
    \begin{tabular}{l |  ccc | ccc | ccc}
      \toprule
      \textbf{Model Name} & \multicolumn{3}{c|}{\textbf{MATH500}} & \multicolumn{3}{c|}{\textbf{AIME24}} & \multicolumn{3}{c}{\textbf{AIME25}} \\
      \cmidrule(lr){2-10}
      & ACC & LEN & NAG $\downarrow $ & ACC & LEN & NAG $\downarrow $ & ACC & LEN & NAG $\downarrow $  \\
      \midrule
      Base Model & 91.3 & 3701 & -- & 50.6 & 10382 & -- & 36.7 & 10958 & -- \\
      EA with $\eta$ = 0.1 & 91.7 & 2170 & -1.36 & 53.5 & 7543 & -10.96 & 37.1 & 8327 & -2.22 \\
      EA with $\eta$ = 0.2 & 91.2 & 1870 & 0.16 & 50.6 & 7172 & 0.0 & 36.9 & 8241 & -1.09 \\
      ME with $\eta$ = 0.1 & 92.1 & 2170 & -1.36 & 53.8 & 6824 & -10.8 & 36.3 & 7311 & 1.89 \\
      ME with $\eta$ = 0.2 & 91.4 & 2003 & -0.16 & 51.7 & 6658 & -3.63 & 36.5 & 7142 & 0.91 \\
      \bottomrule
    \end{tabular}
  }
  \label{tab:ablaiton}
\end{table}
\paragraph{Training Accuracy}
\textbf{Difficulty-aware entropy regularization improves data efficiency.}
Figure \ref{fig:training_acc_comparison} highlights a distinct performance gap: {\ours} (equipped with maximum-entropy loss) consistently surpasses the standard Length-Penalty baseline in training accuracy.
Our approach consistently attains higher training accuracy, which can be attributed to high-entropy exploration on challenging questions, enabling more effective learning from the same data.
Meanwhile, by applying length penalties selectively based on historical question accuracy, our approach extracts a richer learning signal from the same number of rollouts, demonstrating significantly improved data efficiency and a more robust training trajectory compared to the uniform penalty strategy.


\subsection{Ablation Study}
In this subsection, we conduct ablation studies of the proposed method, examining different forms of entropy regularization and the effect of the length-penalty coefficient.

\paragraph{Entropy Regularization Form}
\textbf{Entropy-based advantage allocates more reasoning budget to intractable questions.}
In Table \ref{tab:main_results}, we observe that {\ours}\textbf{-EA} produces longer average responses than {\ours}\textbf{-ME
}.
To better understand this difference, we analyze by examining the distribution of response lengths on AMC23, as shown in Figure \ref{fig:token_distribution_comparison}.
The EA variant tends to allocate more generation budget to difficult questions, sometimes reaching the maximum response length on instances it still fails to solve.
This behavior stems in part from intrinsic tendency toward length expansion of entropy-based advantage technique \cite{cheng2025reasoning} and in part from its explicit encouragement of high-entropy token generation.
As a result, the model assigns higher probability mass to connective tokens that is critical for reasoning and  facilitates reflection, which can prolong inference trajectories.

\paragraph{The coefficient of Entropy Regularization}
\textbf{Strong length penalties can suppress the benefits of entropy regularization, yet accuracy remains comparable to the base model.}
We conduct an ablation study on the length-penalty coefficient. Intuitively, a larger length penalty will further compress the reasoning process, but also risk degrading accuracy. As shown in Figure \ref{fig:length_comparison}, for both entropy-regularization variants, increasing the length penalty consistently yields shorter reasoning on the training set. However, Table \ref{tab:ablaiton} indicates that under EA, responses to challenging questions still remain relatively long even with larger penalties, consistent with our earlier analysis that EA allocates more budget to hard instances. 
Consequently, although aggressive length penalties induce a marginal decline in accuracy compared to milder settings, the performance does not collapse.
Instead, it remains comparable to the base model, suggesting that our method successfully guides the model to prune redundant ``fluff'' tokens while retaining the critical reasoning steps necessary for correct deduction.

%% file: section/related-work.tex
\section{Related Work}

\paragraph{Reasoning Compression}

The substantial inference cost associated with excessive tokens and redundant intermediate reasoning has motivated recent work on compressing reasoning processes.
In the earlier stage, several approaches leverage the model’s own output as supervision signals to compress the reasoning process \citep{munkhbat2025self, chen2024not, xia2025tokenskip, huang2025reasoning, xie2}.
\citet{chen2024not} samples multiple responses generated by the model itself to perform DPO \citep{rafailov2023dpo}, allowing the learning of length-aware preferences.
FS-BoN \citep{munkhbat2025self} leverages self-generated concise reasoning paths obtained by best-of-N sampling for subsequent fine-tuning.
SEER \citep{huang2025reasoning} generates CoT rationales and answers, discards incorrect outputs, and fine-tunes iteratively on the correct rationale–answer pairs, reaching reasoning performance comparable to models that are thirty times larger.
TokenSkip \citep{xia2025tokenskip} employs another LLM to estimate the semantic importance of individual CoT tokens, and select tokens of high importance to generate compressed training data.
More recently, RL has been increasingly used to explicitly control reasoning length in LLMs.
These approaches typically rely on multi-stage training paradigms or the incorporation of length-aware reward signals.
Thinkless \citep{fang2025when} constructs a dataset of short-term and long-term reasoning trajectories distinguished by designated control tokens, enabling the model to learn token-based switches that activate different reasoning modes.
AutoThink \citep{tu2025autothink} progressively refines reasoning strategies through staged reward shaping, achieving favorable accuracy–efficiency
trade-offs.
More generally, incorporating length-related penalty terms into the RL objective has become a common design choice.
FEDH \citep{ling2025fedh} and DR. SAF \citep{chen2025aware} rely on human-defined length priors to encourage compressed reasoning, while Length-Penalty \citep{arora2025length} and LC-R1 \citep{cheng2025LC-R1} design adaptive length penalties based on inter-group
length comparisons among the generated outputs.
These methods are often constrained by manually specified length priors or struggle to balance accuracy and reasoning length, which limits their scalability. 
In contrast, {\ours} adaptively compresses reasoning within a single training phase while preserving the performance.

\paragraph{Entropy Regularization}

Recently, several studies have investigated the phenomenon of entropy collapse in reinforcement learning with verifiable rewards (RLVR) \citep{yu2025dapo, jiang2025rethinking, cui2025entropy, park2025clip}.
A key empirical finding is the trade-off between policy entropy and task performance \citep{cui2025entropy}, which has motivated renewed interest in entropy regularization as a mechanism for stabilizing training and improving generalization.
Subsequent work has primarily focused on dynamically modulating policy entropy \citep{cui2025entropy, wang2025arbitrary, jiang2025rethinking, a3po} during RL learning progress to preserve exploratory behavior and improve model performance.
Representative methods include unbalanced clipping of positive and negative advantages \citep{yu2025dapo, park2025clip, zhu2025surprising}, decoupled optimization strategies for high-entropy tokens \citep{cao2025efficient, jiang2025rethinking, li2025cure}, and the explicit optimization of entropy-related advantage terms \citep{zhang2025edge, cheng2025reasoning}.
Recent analyses of positive and negative samples \cite{zhu2025surprising, a3po} provide an informative perspective on why reasoning compression is challenging for strong LLMs like R1-distill-Qwen-2.5-7B. In particular, optimizing on positive samples tends to reduce policy entropy \cite{zhu2025surprising}, and token-level schemes that emphasize high-entropy tokens in positive samples or low-entropy tokens in negative samples can still drive entropy downward \cite{a3po}. In strong LLMs, these effects are amplified by high baseline accuracy and confidence, which make the on-policy distribution increasingly peaked during RL updates. Consequently, RL-based length optimization on strong models often exhibits a pronounced trade-off between accuracy and response length, accompanied by rapid decay of the policy entropy.
Moreover, prior studies have demonstrated that maintaining an appropriate level of entropy can substantially improve Pass@$k$ performance \citep{wang2025beyond, zhu2025surprising, chen2025pass}, suggesting a deeper impact on the model’s reasoning ability \citep{yue2025does}.
However, a recurring empirical observation in these works is that response lengths tend to increase as performance improves, particularly when additional entropy regularization is introduced \citep{wang2025beyond, cui2025entropy, cheng2025reasoning}, which conflicts with our objective of compressing the reasoning process. 
To resolve this tension, we selectively apply entropy regularization at the question level based on model-dependent difficulty, preserving model performance while explicitly biasing
learning toward shorter and more compact reasoning trajectories.

%% file: section/conclusion.tex
\section{Conclusion}

This work identifies entropy collapse as a central obstacle in RL-based reasoning compression: aggressively optimizing for brevity shrinks the effective exploration space, which is particularly harmful for difficult questions that require diverse intermediate hypotheses or alternative reasoning paths. Consequently, this collapse limits the benefits of RL and induces accuracy degradation under tighter length budgets.
To address this failure mode, we introduce {\ours}, which explicitly separates where to compress and where to explore. The key idea is to preserve exploration on questions that are currently hard for the model via difficulty-aware selective entropy regularization, while allowing easy questions to be compressed more aggressively.
{\ours} further stabilizes length control with a question-level dynamic optimal-length penalty that anchors compression to the historically shortest correct trajectory for each question. By penalizing correct responses relative to this per-question reference, the length signal remains stable even as response-length distributions shift over training and remains effective when entropy regularization temporarily inflates length on hard instances. Across six reasoning benchmarks, {\ours} consistently reduces response length while maintaining accuracy comparable to the base model, and it improves Pass@$k$ relative to length-only optimization, which supports the claim that principled entropy control during compression preserves genuine reasoning capability rather than merely amplifying confidence.

\section*{Acknowledgments}
 
This work was supported by the National Natural Science Foundation of China (Nos.62506166, U2441285),  the Natural Science Foundation
of Jiangsu Province (No.BK20251365), the China Postdoctoral Science Foundation (No. 2025M774283), the Scientific Research Starting Foundation of
Nanjing University of Aeronautics and Astronautics (No.1015-YAH24096), and the High Performance
Computing Platform of Nanjing University of Aeronautics and Astronautics.
This research is also sponsored by the DiDi GAIA Collaborative Research Funds (No. CCF-DiDi GAIA202507) and CAAI-MindSpore Open Fund (CAAIXSJLJJ 2025 MindSpore 01), developed on OpenI Community.

%% file: section/appendix.tex
\appendix

\section{Experimental Setup}
We utilize the verl framework \cite{sheng2024hybridflow} for RL training.
To reduce memory overhead, we fine-tune the model with LoRA \cite{Lora} on two nodes (16 A100 GPUs in total). 
Table \ref{tab:training} reports the training hyper-parameters, with the specific settings highlighted in bold.
\begin{table}[ht]
    \centering
    \begin{tabular}{l|c}
        \toprule
         Parameter Name & Value \\
         \midrule
         advantage estimator & grpo \\
         training batch size & 512 \\
         max prompt length & 2048  \\
         max response length & 20480 \\
         ppo mini batch size  & 32 \\
         ppo micro batch size per gpu  & 2 \\
         log prob micro batch size per gpu & 4 \\
         use kl loss & True \\
         kl loss coefficient & 0.001 \\
         entropy coeff & \textbf{0.001 for ME (0 for EA)} \\
         model parallel size & 2 \\
         gpu utilization & 0.8 \\
         rollout num per question & \textbf{12} \\
         \textbf{lora rank} & \textbf{32} \\
         \textbf{lora alpha} & \textbf{32} \\
         clip ratio high & \textbf{0.28} \\
         learning rate & \textbf{3e-5} \\
         temperature & \textbf{0.6} \\
         \bottomrule
    \end{tabular}
    \caption{Main Parameters of the VERL Training Framework}
    \label{tab:training}
\end{table}

Similarly, we utilize the verl framework for validation experiments without specific settings and report the validation-related hyper-parameters in Table \ref{tab:validation}.

\begin{table}[ht]
    \centering
    \begin{tabular}{l|c}
         \toprule
         Parameter Name & Value \\
         \midrule
         validation batch size & 256 \\
         validation temperature & 0.6 \\
         $\text{top\_p}$ & 0.95 \\
         rollout num per question & 16 \\
         max prompt length & 2048  \\
         max response length & 16000 \\
         model parallel size & 2 \\
         gpu utilization & 0.9 \\
         \bottomrule
    \end{tabular}
    \caption{Validation Parameters of the VERL Training Framework}
    \label{tab:validation}
\end{table}

\section{Additional Ablation Study}
We also conducted ablation studies on the entropy multiplier used in the ME method, the EMA rates of the asymmetric EMA mechanism, and model families beyond Qwen.

Table \ref{tab:entropy_multiplier} reports the model performance under different values of the entropy multiplier. We adopt a 5× entropy multiplier for high-difficulty questions because prior work \citep{cui2025entropy} suggests that this setting can help ensure training stability. Therefore, this choice is primarily empirical and relatively conservative.

\begin{table}[ht]
  \centering
  \caption{Ablation results with different entropy multipliers.}
  \label{tab:entropy_multiplier}
  \resizebox{\columnwidth}{!}{
    \begin{tabular}{lcc|cc|cc|cc|cc|cc}
      \toprule
      \textbf{Model Name} 
      & \multicolumn{2}{c|}{\textbf{GSM8K}} 
      & \multicolumn{2}{c|}{\textbf{MATH500}} 
      & \multicolumn{2}{c|}{\textbf{AIME24}} 
      & \multicolumn{2}{c|}{\textbf{AMC}} 
      & \multicolumn{2}{c|}{\textbf{OlymBench}} 
      & \multicolumn{2}{c}{\textbf{AIME25}} \\
      \cmidrule(lr){2-13}
      & ACC & LEN 
      & ACC & LEN 
      & ACC & LEN 
      & ACC & LEN 
      & ACC & LEN 
      & ACC & LEN \\
      \midrule
      Ours (5$\times$) 
      & 85.8 & 997  
      & 83.6 & 2497 
      & 29.8 & 7713 
      & 72.3 & 4044 
      & 44.7 & 4693 
      & 24.0 & 7350 \\

      2$\times$ 
      & 85.7 & 946  
      & 83.7 & 2454 
      & 29.4 & 7480 
      & 73.0 & 3924 
      & 44.1 & 4864 
      & 23.2 & 7137 \\

      10$\times$ 
      & 86.2 & 1079 
      & 84.1 & 2827 
      & 31.5 & 7845 
      & 73.3 & 4480 
      & 44.9 & 4797 
      & 24.8 & 7596 \\
      \bottomrule
    \end{tabular}
  }
\end{table}

Table \ref{tab:ema_ablation} reports the model performance under different asymmetric EMA rates. Our goal is not to identify the optimal hyperparameter setting. Instead, we adopt asymmetric EMA rates mainly to make the historical accuracy estimate more robust to occasional short-term drops in accuracy during training. This design is intended to improve the stability of difficulty tracking.

\begin{table}[ht]
  \centering
  \caption{Ablation results with different EMA rates.}
  \label{tab:ema_ablation}
  \resizebox{\columnwidth}{!}{
    \begin{tabular}{lcc|cc|cc|cc|cc|cc}
      \toprule
      \textbf{EMA} 
      & \multicolumn{2}{c|}{\textbf{GSM8K}} 
      & \multicolumn{2}{c|}{\textbf{MATH500}} 
      & \multicolumn{2}{c|}{\textbf{AIME24}} 
      & \multicolumn{2}{c|}{\textbf{AMC}} 
      & \multicolumn{2}{c|}{\textbf{OlymBench}} 
      & \multicolumn{2}{c}{\textbf{AIME25}} \\
      \cmidrule(lr){2-13}
      & ACC & LEN 
      & ACC & LEN 
      & ACC & LEN 
      & ACC & LEN 
      & ACC & LEN 
      & ACC & LEN \\
      \midrule
      Ours (0.2, 0.05) 
      & 85.8 & 997  
      & 83.6 & 2497 
      & 29.8 & 7713 
      & 72.3 & 4044 
      & 44.7 & 4693 
      & 24.0 & 7350 \\

      (0.1, 0.01) 
      & 85.6 & 1099 
      & 84.2 & 2511 
      & 28.8 & 7704 
      & 75.2 & 4654 
      & 45.0 & 4653 
      & 23.1 & 7862 \\

      (0.4, 0.1) 
      & 85.7 & 1066 
      & 84.5 & 2323 
      & 29.6 & 7893 
      & 71.4 & 4774 
      & 44.0 & 5058 
      & 20.8 & 7996 \\
      \bottomrule
    \end{tabular}
  }
\end{table}

We also conduct experiments on Llama-3.2-3B-Instruct, and the results are shown in the table \ref{tab:llama_ablation}. Notably, the R1-Distill series is obtained by distilling much stronger teacher models, so its headroom for further improvement through RL is naturally limited. In contrast, for a non-distilled model, RL could yield much more substantial gains.

\begin{table}[ht]
  \centering
  \caption{Ablation results on Llama-3.2-3B-Instruct.}
  \label{tab:llama_ablation}
  \resizebox{\columnwidth}{!}{
    \begin{tabular}{lcc|cc|cc|cc}
      \toprule
      \textbf{Llama-3.2-3B-Instruct} 
      & \multicolumn{2}{c|}{\textbf{GSM8K}} 
      & \multicolumn{2}{c|}{\textbf{MATH500}} 
      & \multicolumn{2}{c|}{\textbf{AMC}} 
      & \multicolumn{2}{c}{\textbf{OlymBench}} \\
      \cmidrule(lr){2-9}
      & ACC & LEN 
      & ACC & LEN 
      & ACC & LEN 
      & ACC & LEN \\
      \midrule
      Base
      & 11.8 & 219
      & 27.5 & 1321
      & 15.5 & 2650
      & 9.6 & 2504 \\

      Ours (ME)
      & 77.5 & 195
      & 43.0 & 274
      & 20.6 & 268
      & 11.0 & 246 \\
      \bottomrule
    \end{tabular}
  }
\end{table}

\section{Generalization to Code Generation}
We evaluate the trained models on code generation tasks to demonstrate the effectiveness of our method for cross-task reasoning compression.

We evaluate the trained models reported in our paper on the code generation task using the HumanEval+ benchmark \citep{Humaneval}. The results in Table \ref{tab:code_generation} show that our method enables more efficient and more accurate reasoning across tasks. This is consistent with our goal of achieving reasoning compression while maintaining or further improving the reasoning capability of the original base model.

\begin{table}[ht]
  \centering
  \caption{Results on code generation benchmarks.}
  \label{tab:code_generation}
  \begin{tabular}{lcccc}
    \toprule
    \textbf{Model Name} 
    & \textbf{avg@16} 
    & \textbf{$\Delta \uparrow$} 
    & \textbf{Length} 
    & \textbf{$\Delta \downarrow$} \\
    \midrule
    R1-Distill-Qwen2.5-7B 
    & 32.5 & / 
    & 676.5 & / \\

    Ours (7B-ME) 
    & 36.5 & +4.0 
    & 626.0 & -50.5 \\

    R1-Distill-Qwen2.5-1.5B 
    & 34.7 & / 
    & 645.1 & / \\

    Ours (1.5B-EA) 
    & 40.1 & +5.4 
    & 582.0 & -63.1 \\
    \bottomrule
  \end{tabular}
\end{table}